\documentclass[10pt,journal,compsoc]{IEEEtran}

\ifCLASSOPTIONcompsoc
  \usepackage[nocompress]{cite}
\else
  \usepackage{cite}
\fi

\ifCLASSINFOpdf
   \usepackage[pdftex]{graphicx}
\else
   \usepackage[dvips]{graphicx}

\fi
\usepackage{amsmath}
\usepackage{algorithm}
\usepackage{algorithmic}
\usepackage{array}

\ifCLASSOPTIONcompsoc
  \usepackage[caption=false,font=footnotesize,labelfont=sf,textfont=sf]{subfig}
\else
  \usepackage[caption=false,font=footnotesize]{subfig}
\fi
\usepackage{fixltx2e}
\usepackage{stfloats}
\usepackage{url}
\usepackage{color}
\usepackage{bm}
\usepackage{hyperref}
\usepackage{booktabs}
\usepackage{multirow}
\usepackage{makecell}
\usepackage{amsthm,amssymb,amsfonts}
\usepackage{diagbox}
\usepackage{footnote}
\usepackage{mathrsfs}
\usepackage{amsfonts,amsmath,bm}

\begin{document}
\title{Leave Zero Out: Towards a No-Cross-Validation Approach for Model Selection}
\newtheorem{myDef}{Definition}
\newtheorem{myPro}{Problem}
\newtheorem{corollary }{Corollary }
\newtheorem{theorem}{Theorem}
\newtheorem{remark}{Remark}
\newtheorem{corollary}{\bf Corollary}

\author{Weikai~Li, Chuanxing~Geng and Songcan~Chen% <-this % stops a space
\IEEEcompsocitemizethanks{\IEEEcompsocthanksitem The authors are with College of Computer Science and Technology, Nanjing University of Aeronautics and Astronautics of (NUAA), Nanjing, 211106, China.\protect\\
E-mail: \{leeweikai; gengchuanxing; s.chen\}@nuaa.edu.cn.
\IEEEcompsocthanksitem Corresponding author is Songcan Chen.}% <-this % stops an unwanted space
\thanks{Manuscript received April 19, XXXX; revised August 26, XXXX.}}


\markboth{Journal of \LaTeX\ Class Files,~Vol.~14, No.~8, August~2015}%
{Shell \MakeLowercase{\textit{et al.}}: Bare Demo of IEEEtran.cls for Computer Society Journals}



\IEEEtitleabstractindextext{%
\begin{abstract}
As the main workhorse for model selection, Cross Validation (CV) has achieved an empirical success due to its simplicity and intuitiveness. However, despite its ubiquitous role, CV often falls into the following notorious dilemmas. On the one hand, for small data cases, CV suffers a conservatively biased estimation, since some part of the limited data has to hold out for validation. On the other hand, for large data cases, CV tends to be extremely cumbersome, e.g., intolerant time-consuming, due to the repeated training procedures. Naturally, a straightforward ambition for CV is to validate the models with far less computational cost, while making full use of the entire given data-set for training. Thus, instead of holding out the given data, a cheap and theoretically guaranteed auxiliary/augmented validation is derived strategically in this paper. Such an embarrassingly simple strategy only needs to train models on the entire given data-set once, making the model-selection considerably efficient. In addition, the proposed validation approach is suitable for a wide range of learning settings due to the independence of both augmentation and out-of-sample estimation on learning process.  In the end, we demonstrate the accuracy and computational benefits of our method by extensive evaluation on multiple data-sets.
\end{abstract}

}

\maketitle

\IEEEdisplaynontitleabstractindextext

% \fi
\IEEEpeerreviewmaketitle

\begin{figure}[t]
    \centering
    \subfloat[The main idea behind CV: CV estimates the generalization error by the out-of-sample estimation, the entire data-set has to split for training and validation multiple times. ]{
        \includegraphics[width=0.5\textwidth]{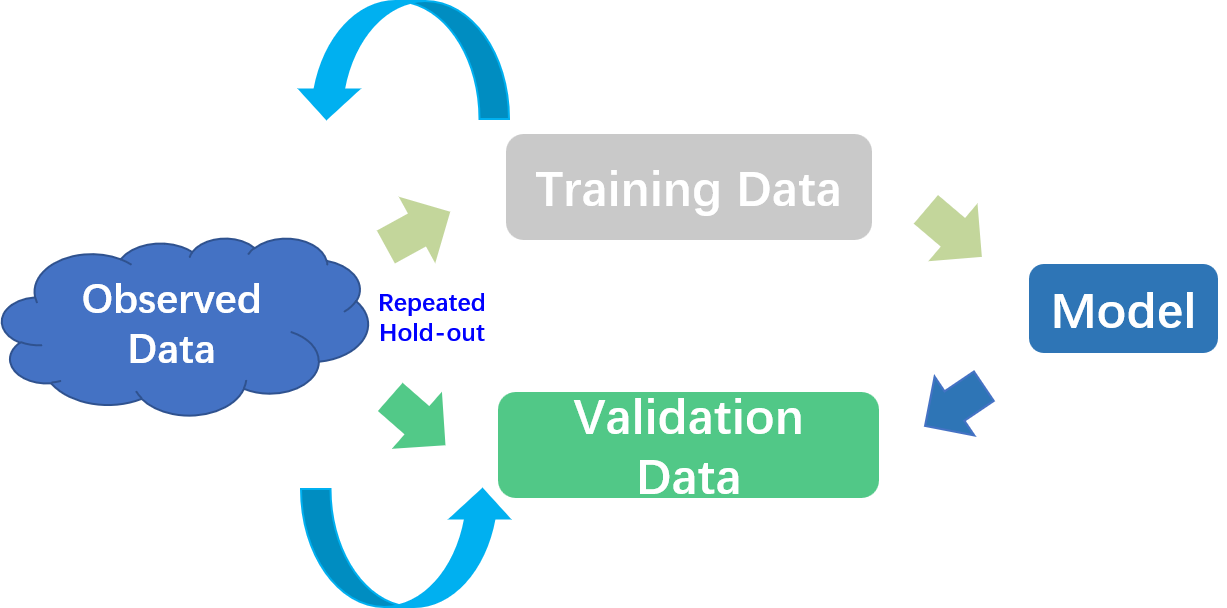}
        \label{figure1a}
    }\\
    \subfloat[The main idea behind LZO: The model is trained on the entire given data-set only once and validated on the cheap auxiliary validation data-set. ]{
        \includegraphics[width=0.5\textwidth]{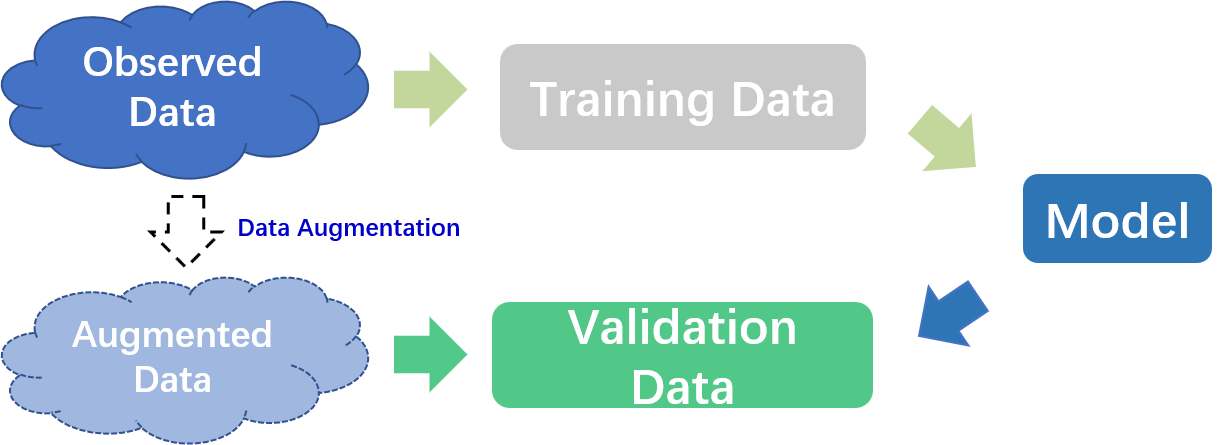}
        \label{figure1b}
    }
    \caption{The main idea behind (a)CV, (b) LZO.}
    %\label{fig:fig_micPerMon}
\end{figure}

\IEEEraisesectionheading{\section{Introduction}\label{sec:introduction}}
\IEEEPARstart{C}{ross} Validation (CV) is undeniably the most commonly-used model selection strategy in the machine learning field \cite{arlot2010survey}. The main idea behind CV is the out-of-sample estimation through hold-out or data splitting \cite{austern2020asymptotics,cawley2004fast} as shown in Figure \ref{figure1a}, since the generalization error is not directly computable  \cite{geisser1975predictive,stone1974cross,shao1997asymptotic,celisse2008model,arlot2011segmentation}. Specifically, part of data (forming the validation set) is held out to evaluate the performance of the candidate models, while the remainder (forming the training set) is left for training. Commonly, a single hold-out yields a validation estimate of the risk, and the averaging over several hold-outs yields a cross-validation estimate. Due to the nature of the out-of-sample estimation, compared with the resubstitution error \cite{kohavi1995study}, CV effectively avoids over-fitting.

Despite its empirical success, CV is vulnerable to the following two obstacles due to the hold-out or data splitting: Firstly, according to a series of statistical learning theories \cite{valiant1984theory,vapnik1999overview,vapnik2013nature}, its estimates tend to be conservatively biased, especially for the small scale case, which has already been empirically verified \cite{tsamardinos2018bootstrapping,rad2018scalable}; Secondly, CV commonly requires a repeated model training procedure, resulting in a severe computational demand, which is especially intolerable for the large-scale model selection. Very naturally, one straightforward ambition for CV is to validate the models with much less computational cost while fully utilizing the entire given data. 
%generalized CV 
Towards this ambition, several efforts have been devoted to approximating CV by replacing the most cumbersome model re-training in CV with an inexpensive surrogate. Specifically, for the empirical risk minimization based models, a line of researches is proposed to approximate CV via the Newton method \cite{rad2018scalable,beirami2017optimal,wilson2020approximate,ghosh2020approximate,seeger2008cross} or the classical infinitesimal jackknife (IJ) from statistics \cite{giordano2019higher,giordano2019swiss}. In the context of kernel-based models, a series of studies \cite{liu2018fast,liu2019efficient,liu2019efficient} approximates CV by representing the Bouligand Influence Function (BIF) \cite{christmann2008bouligand} as the terms of Taylor expansions. For the deep-learning-based models, Corneanu, \MakeLowercase{\textit{et al.}} \cite{corneanu2020computing} directly utilizes the persistent topology measures \cite{corneanu2019does} to estimate the performance on the unseen testing data-set. For linear-fitting-based models under the squared-error loss, generalized cross-validation provides a convenient approximation to leave-one-out cross-validation based on the trace of the smoothing matrix \cite{craven1978smoothing}. Although the methods mentioned above can greatly reduce the computational cost of CV, most of them can only work on a single specific type of model, which limits their scope of applications. Besides, such approaches train and evaluate the models on the same data-set, easily yielding an overoptimistic estimate \cite{larson1931shrinkage}, which further limits their performance. 

On the contrary, this paper employs again the idea of the out-of-sample estimation \cite{mosteller1968data}, where we aim to obtain a validation approach that is not only efficient for computation, but also effective for validation and easy for application. To this end, instead of the existing commonly-used validation methods such as approximate CV or the hold-outs, we strategically derive an extra cheap auxiliary/augmented validation set directly from the given data-set via data augmentation (DA) \cite{van2001art,ratner2017learning,cubuk2019autoaugment,zhao2019data,wu2020generalization} as shown in figure \ref{figure1b}. Apparently, the augmented validation set plays a key role for selecting the ideal model. Fortunately, we can easily find a practical augmentation strategy whose principle is quite mild. Further, based on the Janson-Shannon (JS) divergence, we provide a theoretical upper bound of the estimation bias to confirm its rationality. %Apparently, the estimation via the augmented validation set plays a key role for selecting the ideal model. To this end, With the guidance of the Janson-Shannon (JS) divergence, we provided a theoretical  upper bound of the estimation bias. Fortunately, from the theoretical results, we can easily derive a practical augmentation strategy whose principle is quite mild.
%\textcolor{red}{x}
%Apparently, augmented validation set plays a key role for selecting the positive model. Fortunately, we can easily find a practical augmentation strategy whose principle is quite mild. Further, with the guidance of the Janson-Shannon (JS) divergence, we provided a theoretical  upper bound of the estimation bias to confirm its rationality.

It is worth pointing out that such an operation does not require to leave even one data out. Thus, we name the proposed scheme as Leave-Zero-Out (LZO), following the naming of Leave-One-Out (LOO) in the traditional CV \cite{allen1974relationship,geisser1975predictive}. Compared with the traditional CV, LZO just needs one-time training, hence can significantly improve the efficiency of validation. Moreover, the estimation of LZO can be least biased, since LZO directly estimates the performance on the final returned model. Meanwhile, it establishes a desired model whose performance is potentially superior, since LZO makes full use of the whole precious training data-set. Likewise, it is especially suitable for the small size data-sets. Obviously, such characteristics make the proposed LZO NOT limited to the supervised learning, while also applicable to some challenging learning settings with quite limited labeled data, e.g., semi-supervised learning.  %副词修饰形容词

To validate the efficiency and the effectiveness of LZO, we conduct multiple experiments on 20 supervised data-sets and 6 semi-supervised data-sets. The results demonstrate that LZO gains not only comparable accuracy to the traditional CV on the supervised learning setting while much accuracy improvement on the semi-supervised learning setting, but also significant improvement in efficiency. To facilitate the repetition of our work, our code is released at GitHub \footnote{\href{https://github.com/Cavin-Lee/LZOV}{https://github.com/Cavin-Lee/LZOV}.}. In summary, our contributions can be high-lighted as follows:

\begin{enumerate}[\IEEEsetlabelwidth{8)}]
\item We develop an embarrassingly \emph{simple} and \emph{efficient} validation approach named LZO. It is also \emph{general} due to the independence of both augmentation and out-of-sample estimation on learning process. LZO offers a new paradigm for practical model selection.
\item We provide an \emph{almost-free} data augmentation practical principle for generating the auxiliary validation set.
\item We demonstrate the \emph{effectiveness} of the proposed LZO approach by a thorough evaluation on several data-sets and models.
\end{enumerate}

The remainder of this paper is organized as follows. First, we briefly review some related works in Section \ref{section2}. Then, in Section \ref{section3}, we present the details of the proposed LZO for model selection. Next, in Section \ref{section5}, we validate the performance of the proposed LZO approach compared with standard CV procedure. Finally, we conclude the entire paper in Section \ref{section6}.

%\begin{figure*}[!t]
%\centering
%\subfloat[Distribution of the original data set.]{\includegraphics[width=2.2in]{fig2.pdf}%
%\hfil
%\subfloat[Traditional recognition/classification problem.]{\includegraphics[width=2.25in]{fig3.pdf}%
%\label{fig_second_case}}
%\hfil
%\subfloat[Open set recognition/classification problem.]{\includegraphics[width=2.21in]{fig4.pdf}%
%\label{fig_third_case}}
%caption{Comparison of between traditional classification and open set recognition. Fig. 2(a) denotes the distribution of original dataset including KKCs 1,2,3,4 and UUCs ?5,?6, where KKCs appear during training and testing while UUCs may appear or not during testing. Fig. 2(b) shows the decision boundary of each class obtained by traditional classification methods, and it will obviously misclassify when UUCs ?5,?6 appear during testing. Fig. 2(c) describes open set recognition, where the decision boundaries limit the scope of KKCs 1,2,3,4, reserving space for UUCs ?5,?6. Via these decision boundaries, the samples from some UUCs are labeled as "unknown" or rejected rather than misclassified as KKCs.}
%\label{fig_sim}
%\end{figure*}

\section{Related Works}\label{section2}

In this section, we briefly present the most related works, and highlight the differences with the proposed LZO.
\subsection{Cross Validation}
This study aims to not only accelerate the validation procedure with effectiveness but also provide a CV-free strategy for model selection. As one of the gold standards \cite{arlot2010survey,ghosh2020approximate} for model selection, CV can effectively help to select the model by a simple data splitting mechanism under a basic intuition of the out-of-sample estimation \cite{stone1974cross,larson1931shrinkage,geisser1975predictive}, as shown in Figure \ref{figure1a}. According to the hold-out or data splitting strategies, the standard CV can be summarized into following two paradigms:

\noindent
\textbf{Exhaustive Data Splitting:} Such approach is also known as the Leave-$p$-Out (LPO) CV \cite{shao1993linear,ronchetti1997robust}, every possible subset of $p$ data is successively held out and used for validation. Note that LPO with $p=1$ turns to the most classical exhaustive Leave-One-Out CV \cite{kearns1999algorithmic}. Though the estimation bias of this approach should be the small, the computational cost is intolerable. Also, it may collapse in the sense that it can provide extremely misleading estimates in degenerate situations, especially when $p$ is small \cite{witten2002data}.

\noindent
\textbf{Partial Data Splitting:} Considering the number of exhaustive data splitting can be computationally intractable, even when $p$ is small, partial data splitting schemes have been proposed as alternatives. Only part of the appropriate subset of the given data-set is held out in this paradigm, such as $K$-fold CV \cite{rodriguez2009sensitivity,zhang1993model}, Balanced Incomplete CV (BICV) \cite{john1998statistical}, Repeated learning-testing (RLT) \cite{bowman1984alternative} and Monte-Carlo CV (MCCV) \cite{picard1984cross}. Though such a strategy leads to speed-up in CV, its estimation commonly suffers a large bias \cite{tsamardinos2018bootstrapping,rad2018scalable}. 

Notably, counting the number of models created by traditional CV, let $C$ be the number of candidate learning methods, and $K$ be the repeated times. To produce the final model, CV creates $K \times C$ models and once the best configuration is picked, one more model will be produced, leading to $K \times C + 1$ models for final model production. As a result, such approach entails solving multiple expensive model training procedures, which makes the validation procedure extremely cumbersome. Further, the estimates of traditional CV are conservatively biased, while the final model is re-trained on the entire given data-set for practical applicability but the estimates are produced by the models trained on the part of given data-set. 

In contrast, although still tackling the model selection via the out-of-sample estimation, with the help of auxiliary validation set, LZO is able to produce the final model by training on the entire given data-set and only need once training. This makes LZO both effective and efficient for model selection. Note that there is also some studies on the approximation of the general CV for efficient model selection \cite{beirami2017optimal,wilson2020approximate,ghosh2020approximate,liu2018fast,craven1978smoothing}. However, almost all of them are specially designed for the specific type of models, naturally, limiting their generality to great extent. Thus, we do not consider these approximation methods in this paper.

%  要不要加DA的定义

\section{Leave-Zero-Out Validation} \label{section3}

In this section, towards a CV-free model selection, we provide a novel validation approach named Leave-Zero-Out (LZO).  Based on DA trick, albeit without hold-out or data-splitting, LZO remains tackling the model selection by using the most simple and intuitive out-of-sample estimation. In the following, we first introduce some preliminaries and notations for better representing the LZO. Then, we provide a brief motivation and a sketch pipeline of the LZO strategy. Next, to confirm its rationality, we theoretically analyze the upper bound of the estimation bias based on the Janson-Shannon (JS) divergence. In the end, with the guidance of the theoretical results, we derive a practical principle of DA.
\subsection{Preliminaries and Notations} 
%\subsection{Notations}
Here, we briefly clarify the notations used in the rest of this paper. Let $\mathcal{X} \in \mathbb{R}^d$ and $\mathcal{Y} \in \mathbb{R}$ be the input and output space, respectively. Then, we consider the given data-set $\mathcal{D}=\{(\mathbf{x}_i,y_i)\}_{i=1}^n$ with $n$ samples, which is drawn from the unknown distribution $\mathbb{P}$, where $\mathbf{x}_i \in \mathcal{X}$  and $y_i \in \mathcal{Y}$ are the input feature and the output of $i$-th sample, respectively. It should be noted that the output of some samples might be unknown, e.g., in semi-supervised learning \cite{zhu2009introduction,chapelle2009semi} or unsupervised learning \cite{barlow1989unsupervised}). Let $\mathcal{V}=\{(\mathbf{x}_i^v,y_i^v )\}_{i=1}^m$ be the auxiliary/augmented validation data-set with $m$ samples from the unknown distribution $\mathbb{Q}$. Denote $Q=\{\mathit{f}_i \}_{i=1}^C$ as the candidate configuration set where $\mathit{f}_i:\mathcal{D} \rightarrow \mathcal{H}$ is the $i$-th learning configuration, $\mathcal{H}$ is the hypothesis space with VC dimension $v$ and $C$ is the number of the candidate configurations. Denote $\mathcal{A}:\mathcal{D}\rightarrow\mathcal{V}$ as an operator for data augmentation. We define $\ell$ as the appropriate loss function within an interval $B = \mathop{\max}(\ell)- \mathop{\min}(\ell)$. The definitions of all notations are shown in Table \ref{table1}.
\begin{table}[]
    \centering%\normalsize
    \caption{Some Definitions of Variables}
    \label{table1}
\begin{tabular}{c|l}
\hline\hline

    Notation                     & Description \\ \hline
    $\mathcal{D}$,$\mathcal{V}$  & given/auxiliary data-set \\
    $n$,$m$                      & number of given/auxiliary samples \\
    $\mathbb{P}$,$\mathbb{Q}$    & joint distribution of training/validation set\\
    $\mathbf{x}$                 & input feature\\
    $y$                          & output \\
    $Q$                          & candidate configuration set\\
    $\mathit{f}$                 & learning configuration\\
    $C$                          & number of candidate configurations\\
    $G$                          & learned model\\
    $\mathcal{A}$                & augmentation operator\\
    $\ell$                       & appropriate loss function\\
    $\mathcal{H}$                & hypothesis space\\
    $v$                          & VC dimension\\
    $\mathcal{L}(G,\mathcal{D})$ & empirical risk of model $G$ on the data-set $\mathcal{D}$\\
    $\mathcal{L}_\mathbb{P}(G)$  & expected risk of model $G$ on the distribution $\mathbb{P}$\\ \hline\hline
\end{tabular}
\end{table}

\subsection{Motivation}
As mentioned above, the most cumbersome pipelines of the traditional CV are the repeated hold-out and the expensive model re-training procedures. Thus, a straightforward ambition is to validate the model performance without data-splitting or hold-out, while training the models on the entire given data-set and only once. To achieve this, a quite intuitive motivation is to generate the cheap auxiliary/augmented data-set by an operator $\mathcal{A}$ for validation, whose main idea is illustrated in Figure \ref{figure1b}. Specifically, the entire validation strategy of LZO is given in ALGORITHM \ref{alg2}. In this way, the performance of the model, i.e., $\mathit{f}_i(\mathcal{D})$ is validated on the augmented validation set. Also, it can make full use of the whole precious given data-set and further establish a desired model whose performance is potentially superior. Moreover, to produce a final deployed model, for each learning configuration, LZO needs only one-time training, and once the best configuration is determined, no more model requires to be produced yet, thus leading nearly $K$ times speed-up than standard CV. %In summary,the superiority of this strategy is three-fold intuitively:\\
%\begin{enumerate}[\IEEEsetlabelwidth{8)}]
%\item It is able to train on the entire given data and only once without CV, yielding a quite efficient validation procedure.;
%\item It is able to validate the performance of the final model directly, thus further reducing the estimation bias;
%\item It can make full use of the whole precious given data-set, thus can establish a desired model whose performance is potentially superior;
%\item It is general, due to the procedure of data augmentation is independent on the learning process, making the LZO more flexible. 
%\end{enumerate}
% CV 的策略是由于测试和验证满足独立同分布，然而，LZO很难满足这种假设，因此，我们对其进行了理论的分析
%\subsection{Practical Pipeline of LZO}
%After obtaining the generalized validation set, we can easily use the augmented data-set to evaluate the out-of-sample estimation performance of the model \emph{which is trained on the entire data-set} . The practical algorithm of LZO is summarized in ALGORITHM \ref{alg2}. In particular, counting the number of models created by LZO. To produce the final model, for each learning method, LZO only needs to create one models. Once the best configuration is picked, no more model requires produced. Thus, it can lead to nearly $K$ times faster than standard CV for the final model production.
\begin{algorithm}[ht]
    \caption{Leave-Zero-Out Validation}
    \label{alg2} %算法的标签 
    \begin{algorithmic}[1]
    \REQUIRE ~~\\
    Entire Given Data-set $\mathcal{D}$;\\
    Number $m$; \\
    Candidate Learning Method Set $Q=\{\mathit{f}_1,\mathit{f}_2,\cdots,\mathit{f}_C\}$
    \ENSURE ~~\\ 
    Optimal model $G=\mathit{f}_*(\mathcal{D})$
    %\STATE $//$ while  
    \STATE \#\emph{\textbf{Data Augmentation}}
    \STATE $\mathcal{V}\leftarrow\mathcal{A(D)}$
    \STATE \#\emph{\textbf{Model  Validation}}
    \FOR{$i=1$ to $C$}
    %\WHILE {not converge}
    \STATE $G_i\leftarrow\mathit{f}_i(\mathcal{D})$;
    
    \STATE \textbf{Calculating} $\mathcal{L}(G_i,\mathcal{V})$;
    %\STATE // \emph{leave-zero-out validation via $\mathcal{V}$};
    \ENDFOR%\ENDWHILE 
    \STATE \#\emph{\textbf{Model Selection}};
    \STATE $G=G_i$, $\mathit{f}_*=\mathit{f}_i$ \textbf{where} $i = \mathop{\arg\min}_i\mathcal{L}(G_i,\mathcal{V})$
    \STATE \textbf{Return} $G$.
    \end{algorithmic}
\end{algorithm}
% 提出一个简单的方式，发现可行，并且为了更好的实践，进一步分析。先总体框架，然后 augmented data can be shifted in distribution , however, our analysis state that 
%\subsection{Practical Principles for LZO validation}
% 增广数据存在漂移, 难以保证独立同分布.,

% 增广数据集,独立同分布假设,GAN->复杂性高->自动产生的->目标
%GAN 独立同分布 很少,,代价高,模式崩溃用一个代理近似方式.数据的漂移不能满足独立同分布,
%要设计理论保证的  不是整体增广,条件增广.
%传统CV的估计由独立同分布的重采样来保证，然而，
\subsection{Estimation Bias}
 The existing theoretical result illustrates that the estimation bias of the traditional CV mainly depends on the sample size $n$, since the validation set is holding out from the given limited data-set thus hold the independent and identical distribution (\emph{i.i.d}) with $\mathcal{D}$ \cite{devroye2013probabilistic}. However, the augmented data in LZO can be generally shifted in distribution, making it hard to hold the \emph{i.i.d} assumption. %validation set of traditional CV is obtained by hold-out. 模型性能难以保证，然而我们分析说明，。。。
 In this subsection, to investigate its rationality of the mentioned strategy, we theoretically analyze the upper bound of the estimation bias of LZO.
 %\subsection{Preliminaries}
 %我们来分析了gap,由此，提供了一个DA 的实际的策略。
Firstly, to facilitate the presentation of the ideas, we define the following notations. 

\begin{myDef}[Jensen-Shannon (JS) divergence \cite{fuglede2004jensen,nowozin2016f}]
\label{def1}
 Let $\mathbb{P}$ and $\mathbb{Q}$ be two different distributions, then the JS-divergence between $\mathbb{P}$ and $\mathbb{Q}$ is defined as:
\end{myDef}
\begin{equation}
D_{J S}(\mathbb{P} \| \mathbb{Q})=\frac{1}{2}\left[D_{K L}(\mathbb{P} \| \mathbb{M})+D_{K L}(\mathbb{Q} \| \mathbb{M})\right]
\end{equation}
where $\mathbb{M}=\frac{1}{2}×(\mathbb{P}+\mathbb{Q})$ and $D_{K L}$ is the Kullback-Leibler divergence \cite{van2014renyi}.

\begin{myDef}[Expected Risk]
\label{def2}
 Let $\mathbb{P}$ be a distribution over $\mathcal{X}\times\mathcal{Y}$, $G:\mathcal{X}\rightarrow\mathcal{Y}$ be a model or hypothesis and $\ell(G(\mathbf{x}),y)$ be the pre-defined loss function. Then, the expected risk of model $G$ over distribution $\mathbb{P}$ is defined as:
\end{myDef}
\begin{equation}
    \mathcal{L}_{\mathbb{P}}(G)=\int \ell(G(x), y) d \mathbb{P}(x, y)
\end{equation} 

\begin{myDef}[Empirical Risk]
\label{def3}
Let $\mathcal{D}=\{(\mathbf{x}_i,y_i)\}_{i=1}^n$ be the given data-set. Then, the empirical risk of model $G$ on $\mathcal{D}$ is defined as:
\end{myDef}
\begin{equation}
\label{eq3}
\mathcal{L}(G, \mathcal{D})=\frac{1}{n} \sum_{i=1}^{n} \ell\left(G\left(x_{i}\right), y_{i}\right)
\end{equation}

\begin{myPro}[Model Selection]
\label{pro1}
Given a candidate model set $\{\mathit{f}_i(\mathcal{D}) \}_{i=1}^C$, an appropriate loss function $\ell$, and a training data-set $\mathcal{D}=\{(\mathbf{x}_i,y_i)\}_{i=1}^n$ which is drawn from the distribution $\mathbb{P}$. Model Selection aims to select the optimal model $G=\mathit{f}_*(\mathcal{D})$ which has the minimum expectation risk $\mathcal{L}_\mathbb{P} (G)$. 
\end{myPro}

Then, inspired by the theoretical work of Shui, \MakeLowercase{\textit{et al.}} \cite{shui2020beyond}, we give the theoretical upper bound of the estimation bias as follows:

\begin{theorem}
    \label{the1}
     Let $G\in\mathcal{H}$ be the model or the hypothesis learned from the input data $\mathcal{D}$ by configuration $\mathit{f}$ (i.e.,$G=\mathit{f}(\mathcal{D})$). Let $\mathbb{P}$ and $\mathbb{Q}$ be the distribution of training data and augmented validation data, respectively. Then, for any $\delta\in(0,1)$, with probability at least $1-\delta$, the following holds:
\end{theorem}
\begin{equation}
\label{eqt}
\begin{aligned}
\mid \mathcal{L}_{\mathbb{P}}(G) -\mathcal{L}(G, \mathcal{V}) \mid\leq & \frac{B}{\sqrt{2}} \sqrt{D_{J S}(\mathbb{P} \| \mathbb{Q})} \\
+&\sqrt{\frac{4}{m}\left(v \ln \frac{2 e m}{v}+\ln \frac{4}{\delta}\right)}
\end{aligned}
\end{equation}
%One can see in \textbf{Theorem} \ref{the1} that the upper bound of the discrepancy is bounded by the JS-divergence between two distributions. 
where $e$ is the base of the natural logarithm. The proof is given in the Appendix.
% our analysis state that 
Here, we can easily find in Eq.\ref{eqt} that the estimation bias is bounded by the validation sample size $m$ and the JS-divergence between $\mathbb{P}$ and $\mathbb{Q}$. 
\subsection{Practical Principles for Data Augmentation}
%% 条件生成 模式崩塌 为了更好的估计来保证模型验证， key role is the data augmentation.
% 直接对齐很难，所以我们进行细化。
From \textbf{Theorem} \ref{the1}, we can easily find that the augmented validation set plays a key role for estimation. To minimize the estimation bias, a very straightforward principle for practical augmentation strategy is to control JS-divergence between two distributions. In other words, the augmented validation data-set and the given data-set must not be too dissimilar. 

Note that some DA strategies can directly generate the augmented date-set whose JS-divergence between the given data-set is small. However, such DA strategies may cost expensively and perform poorly. For example, a possibly feasible DA approach to effectively generate the validation set is the generative adversarial network (GAN), whose objective is equivalent to minimizing the dual form of JS-divergence \cite{nowozin2016f}. Unfortunately, the training of GAN is time-consuming. Beside, it has been observed that GAN often suffers from a notorious mode collapse issue \cite{salimans2016improved,srivastava2017veegan}. Further, such adversarial approach can only minimize but not eliminate the gap of JS-divergence between the true and the generated data \cite{glorot2011domain}.

Thus, to obtain a more practical DA strategy to generate the augmented validation set, we further decompose the joint JS-divergence in \textbf{Theorem} \ref{the1} into the marginal and the conditional shift upper bounds, based on the information theoretical chain rule \cite{akaike1998information}:
% much easier
\begin{corollary}
\label{col1}
    The upper bound in Theorem \ref{the1} can be further decomposed as:
\end{corollary}
\begin{equation}
    \begin{aligned}
        \mid \mathcal{L}_{\mathbb{P}}(G) -\mathcal{L}(G, \mathcal{V}) \mid\leq & \frac{B}{\sqrt{2}} D \\
        +&\sqrt{\frac{4}{m}\left(v \ln \frac{2 e m}{v}+\ln \frac{4}{\delta}\right)}
    \end{aligned}
\end{equation}
where $D=\sqrt{\mathbb{E}_{y \sim \mathbb{P}(y)} D_{J S}(x \mid y)+\mathbb{E}_{y \sim Q(y)} D_{J S}(x \mid y)}+\sqrt{D_{J S}(y)}$. For more concise representation, we denote $D_{J S}(x \mid y)$ as $D_{J S}(\mathbb{P}(x \mid y) \| \mathbb{Q}(x \mid y))$ and $D_{J S}(y)$  as $D_{J S}(\mathbb{P}(y) \| \mathbb{Q}(y))$).
The proof is given in the Appendix. 

In particular, the \textbf{Corollary} \ref{col1} provides an alternative guidance to generate the validation set. The discrepancy is alternatively controlled by the label marginal divergence and the semantic (feature) conditional distribution divergence, which naturally derives two practice principles for guiding data generation:

\noindent
\textbf{Controlling the Label Marginal Distribution Divergence:} Since labels are usually categorical variables with the finite classes, we can easily control the label marginal divergence (i.e., $D_{J S}(\mathbb{P}(y) \| \mathbb{Q}(y))\rightarrow 0$) with the given labels. 

\noindent
\textbf{Minimizing the Semantic Conditional Distribution Divergence:} When $D_{J S}(\mathbb{P}(y) \| \mathbb{Q}(y))\rightarrow0$, minimizing the semantic conditional distribution divergence (i.e., $D_{J S}(\mathbb{P}(x \mid y) \| \mathbb{Q}(x \mid y))$)) can effectively control the estimation risk. It is worth noting that minimizing the semantic conditional distribution divergence is much easier than directly minimizing the distribution divergence and highly mitigating the mode collapses \cite{mirza2014conditional}.
\begin{remark} Note that the labels of some samples in the given data-set might be unknown. To address this concern, we introduce the pseudo label as the approximation of the real label.
\label{rem1}
\end{remark}

Assisted by these two principles, as a concept demonstration, we follow the insight of mix-up \cite{zhang2017mixup}, and provide a Label Invariant Mix-up strategy to generate the validation set as shown in ALGORITHM \ref{alg1}. Here, we once again emphasize that such a strategy is only a simple attempt, which means that any other label invariant transformations such as geometric transformations \cite{szegedy2015going}, color transformations \cite{chatfield2014return}, information dropping \cite{devries2017improved} or random erasing \cite{krizhevsky2017imagenet,zhong2020random} can also be adopted to generate the validation set, since they can easily satisfy the mentioned principle by controlling the label distribution divergence (i.e., $D_{J S}(\mathbb{P}(y) \| \mathbb{Q}(y))\rightarrow 0$)). Thus, such principle is mild.

\begin{algorithm}[ht]
    \caption{Label Invariant Mix-Up}
    \label{alg1} %算法的标签 
    \begin{algorithmic}[1]
    \REQUIRE ~~\\
    Entire Given Set $\mathcal{D}$\\
    Number $m$ 
    \ENSURE ~~\\ 
    Augmented Validation Set $\mathcal{V}$
    %\STATE $//$ while
     \FOR{$i=1$ to $m$}
    %\WHILE {not converge}
    \STATE selecting $\{(\mathbf{x}_j,y_j),(\mathbf{x}_k,y_k)\}\in\mathcal{D}$, where $y_j=y_k$;
    \STATE \# \emph{control label marginal divergence};
    \STATE $\lambda \sim Beta(\alpha,\alpha)$, where $\alpha\in(0,+\infty)$;
    \STATE \# \emph{follow the standard mix-up};
   % \STATE Update  $\mathbf{Z}_s^k$ and $\mathbf{Z}_t^k$ ;
    \STATE $\mathbf{x}_i^v=\lambda\mathbf{x}_j+(1-\lambda)\mathbf{x}_k$,$y_i^v=y_j$;
    %\STATE \# \emph{control semantic conditional distribution};
    %\STATE \textbf{IF}  $\lambda$  is small \textbf{THEN} increase $\lambda$ by the step size;
    \ENDFOR%\ENDWHILE 
    \STATE \textbf{Return} $\mathcal{V}$.
    \end{algorithmic}
\end{algorithm}

\begin{remark} Different from the standard DA methods which mainly generate the training set for training models \cite{shorten2019survey}, in this paper, we focus on generating the validation set via DA during the validation phase. To the best of our knowledge, this is the first attempt that DA is adopted to generate the validation set for model selection, which essentially expands the application scope of DA in machine learning. Therefore, it has no exaggeration that our work completes DA as a server for the whole learning life span from training to validation processes.%选择那些样本扩充，diss，简单的进行了实验。数据增广主要是用于训练。bossting
\label{rem2}
\end{remark}
%我们的方法不依赖于任何的模型和损失函数，因此是一般的。
\section{Experiments}
\label{section5}
In this section, we empirically validate the performance of the proposed LZO validation approach on both supervised and semi-supervised learning paradigms.
\subsection{Model Selection in the Supervised Learning Paradigm}
To evaluate the efficiency and the effectiveness of the proposed LZO, we first conduct experiments on the supervised learning paradigm. In fact, LZO is applicable for all supervised learning models. Here, as an instance, we use the most popular support vector machine (SVM) with linear kernel as the base model \cite{chang2011libsvm}, which is achieved by the LibSVM toolbox \footnote{\href{https://www.csie.ntu.edu.tw/~cjlin/libsvm/}{https://www.csie.ntu.edu.tw/~cjlin/libsvm/}.}.
\subsubsection{Data Preparation}
We adopte 20 publicly benchmark data-sets from UCI data-set \footnote{\href{https://archive.ics.uci.edu/}{https://archive.ics.uci.edu/}.} including balance, breast, bupa, clever, dim, dna, glass, heart, housing, ionosphere, iris, mushroom, musk, segment, sonar, testSet, vehicle, vote, wine and wpbc. All data-sets are popular for benchmarking supervised learning algorithms. The statistics of the adopted 20 data-sets are listed in Table \ref{table2}.
\begin{table}[]
\centering%\normalsize
    \caption{Statistics of Benchmark Data-sets on the Supervised Learning Setting}
    \label{table2}
\begin{tabular}{cccc}
\hline\hline
           & \#Class & \#Dimensions & \#Data \\ \hline
balance    & 3       & 4            & 625    \\
breast     & 2       & 9            & 277    \\
bupa       & 2       & 6            & 345    \\
clever     & 2       & 13           & 297    \\
dim        & 2       & 14           & 4192   \\
dna        & 3       & 180          & 2000   \\
glass      & 6       & 9            & 214    \\
heart      & 2       & 13           & 303    \\
housing    & 2       & 13           & 506    \\
ionosphere & 2       & 34           & 351    \\
iris       & 3       & 4            & 150    \\
mushroom   & 2       & 22           & 8124   \\
musk       & 2       & 166          & 6598   \\
segment    & 2       & 8            & 768    \\
sonar      & 2       & 60           & 208    \\
testSet    & 2       & 2            & 100    \\
vehicle    & 4       & 18           & 846    \\
vote       & 2       & 16           & 435    \\
wine       & 3       & 13           & 178    \\
wpbc       & 2       & 33           & 198    \\ \hline\hline
\end{tabular}
\end{table}

\subsubsection{Experimental Setting}
Since some of the traditional CV methods (e.g., LOOCV) tend to be intolerable for the computation cost, we only compare the 10-fold CV for a simple verification due to its popularity. Specifically, for each data-set, we run 10-fold and LZO 100 times with data-sets being split randomly (30\% of all the examples for testing and the other 70\% for training). Following the recommended setting from Change and Lin \cite{chang2011libsvm}, we set the candidate regularized hyper-parameters $C\in[2^{-5},2^{-4},\cdots,2^5 ]$ as the configuration set $Q $ for the linear kernel SVM. We use $m=n$ and $m=10\times n$ as the sample size of the augmented validation set. To evaluate the performance of LZO, we use classification accuracy and the computational cost as the measurements to evaluate the performance.
\subsubsection{Experimental Results and Analysis  }

The classification accuracy and the computational cost are listed in Tables \ref{table3}. We also report the counts of the wins (CoWs) \cite{demvsar2006statistical} in Table \ref{table3}. For each training set, we choose the regularized hyper-parameter $C\in\{2^{-5},2^{-4},\cdots,2^5 \}$ of the linear kernel SVM on the training set and evaluate the accuracy for the chosen parameters on the test set. From those results, we can make several observations as follows. On the most of the data-sets, the accuracy of LZO and 10-fold CV is very similar, neither LZO nor 10-fold CV criterion is shown to be significantly better than the other. It can be observed that the accuracy of LZO on balance, breast, dim, housing, and sonar data-sets significantly outperforms 10-fold CV. Such results illustrate that the LZO can potentially establish more superior models, while the estimates of traditional CV tend to be conservatively biased. For the computational time cost, we can easily find that LZO results in significant computational gains, typically achieving a speed-up of 10 (i.e., $K$) times than 10-fold CV. 

 In addition, from Tables \ref{table3}, we can find that the variance of the accuracy is large when setting $n=m$, such instability may be caused by the randomness of DA. Then, with more augmented validation samples, e.g., $n=10 \times m$, the variance is significantly reduced and the accuracy is almost unchanged, while the additional computational consumption is acceptable. Thus, the robustness of the LZO can be enhanced with more augmented validation samples.
%DA 具有随机性，可以发现在DA在n=m的时候方差很大。但是，随着n=10m,可以发现，方差显著降低，LZO性能更加稳定，而精度几乎不变甚至略有提高，同时时间消费可以接受。
%In addition, we interestingly find that although the final accuracy is not significantly improved with $m$ increase, but the estimation variance are significantly decrease while the additional computational cost is acceptable. Thus, the robust of the LZO can be enhanced by generating more validation samples.
\begin{table*}[]
\centering%\normalsize
    \caption{Mean with Standard Errors of Classification Performance on Supervised Learning Setting (Linear Kernel Based SVM).}
    \label{table3}
\begin{tabular}{c|cc||cc||cc}

\hline\hline
           & \multicolumn{2}{c||}{10-fold} & \multicolumn{2}{c||}{LZO ($m=n$)}& \multicolumn{2}{c}{LZO ($m=10\times n$)} \\
           & Accuracy(\%)     & Time(s)          & Accuracy(\%)        & Time(s)        & Accuracy(\%)        & Time(s)       \\ \hline\hline
balance    &$67.73 \pm 0.40$&$11.65 \pm 0.49$&$70.03 \pm 1.17$&$\mathbf{1.27 \pm 0.01}$&$\mathbf{70.61 \pm 0.00}$&$2.75 \pm 1.16$\\
breast     &$65.47 \pm 0.00$&$150.90\pm 5.63$&$72.62 \pm 0.00$&$\mathbf{17.54 \pm 0.07}$&$\mathbf{72.66 \pm 0.00}$&$24.84 \pm 11.29$\\
bupa       &$\mathbf{70.15 \pm 0.60}$&$68.53 \pm 2.94$&$69.51 \pm 0.33$&$\mathbf{6.64 \pm 0.04}$&$69.86 \pm 0.03$&$6.84 \pm 0.11$\\
clever     &$85.08 \pm 1.11$&$106.47\pm 6.60$&$\mathbf{85.32 \pm 0.69}$&$\mathbf{7.94  \pm 0.04}$&$\mathbf{85.40 \pm 0.01}$&$8.05 \pm 0.03$\\
dim        &$77.37 \pm 1.52$&$50.83 \pm 1.23$&$\mathbf{81.25 \pm 0.00}$&$\mathbf{5.22  \pm 0.03}$&$\mathbf{81.25 \pm 0.00}$&$9.94 \pm 0.31$\\
dna        &$92.90 \pm 0.74$&$32.28 \pm11.25$&$93.44 \pm 0.47$&$\mathbf{6.42  \pm 0.40}$&$\mathbf{93.61 \pm 0.01}$&$6.84 \pm 0.71$\\
glass      &$\mathbf{54.04 \pm 2.53}$&$2.43  \pm 0.93$&$49.82 \pm 0.00$&$\mathbf{0.29  \pm 0.00}$&$49.79 \pm 0.00$&$0.45 \pm 0.07$\\
heart      &$\mathbf{81.41 \pm 0.34}$&$26.90 \pm 3.72$&$81.25 \pm 0.72$&$\mathbf{0.06  \pm 0.01}$&$81.14 \pm 0.00$&$1.62 \pm 0.23$\\
housing    &$52.48 \pm 2.42$&$227.86\pm 3.03$&$79.50 \pm 1.30$&$\mathbf{21.63  \pm 0.14}$&$\mathbf{80.61 \pm 0.16}$&$29.81 \pm 4.28$\\
ionosphere &$88.84 \pm 1.03$&$32.90 \pm 8.14$&${89.17 \pm 0.56}$&$\mathbf{11.43 \pm 0.06}$&$\mathbf{89.56 \pm 0.12}$&${15.96 \pm 2.09}$\\
iris       &$94.45 \pm 0.70$&$ 0.03 \pm 0.00$&$94.34 \pm 2.10$&$\mathbf{0.00 \pm 0.00}$&$\mathbf{97.33 \pm 0.32}$&$0.00 \pm 0.40$\\
mushroom   &$\mathbf{93.30 \pm 0.00}$&$20.74 \pm 0.24$&$87.42 \pm 0.00$&$\mathbf{2.99  \pm 0.02}$&$84.69 \pm 0.00$&$6.73 \pm 0.40$\\
musk       &$\mathbf{85.97 \pm 0.00}$&$137.73\pm 3.67$&$\mathbf{85.97 \pm 0.00}$&$\mathbf{14.17 \pm 0.08}$&$\mathbf{86.97 \pm 0.00}$&$20.73 \pm 2.56$\\
segment    &$\mathbf{90.01 \pm 0.93}$&$402.39\pm 0.90$&$87.73 \pm 0.22$&$\mathbf{44.73 \pm 0.23}$&$87.73 \pm 0.00$&$63.35 \pm 7.29$\\
sonar      &$49.69 \pm 1.98$&$0.50  \pm 0.01$&$\mathbf{55.24 \pm 0.00}$&$\mathbf{0.06 \pm 0.00}$&$\mathbf{55.24 \pm 0.00}$&$0.09 \pm 0.00$\\
testSet    &$\mathbf{100   \pm 0   }$&$ 0.02 \pm 0.00$&$\mathbf{  100 \pm 0   }$&$\mathbf{0.00  \pm 0.00}$&$\mathbf{100 \pm 0}$&$0.01 \pm 0.00$\\
vehicle    &$\mathbf{77.94 \pm 0.07}$&$86.39 \pm 1.32$&$75.63 \pm 0.91$&$\mathbf{7.66 \pm 0.03}$&$75.71 \pm 0.00$&$11.06 \pm 1.79$\\
vote       &$91.91 \pm 0.08$&$28.71 \pm 0.02$&$\mathbf{92.85 \pm 0.46}$&$\mathbf{3.02  \pm 0.00}$&$92.74 \pm 0.05$&$4.12 \pm 0.01$\\
wine       &$\mathbf{98.91 \pm 0.41}$&$19.92 \pm 1.24$&$98.65 \pm 1.03$&$\mathbf{2.23  \pm 0.01}$&$97.78 \pm 0.10$&$2.82 \pm 0.67$\\
wpbc       &$76.34 \pm 0.93$&$23.50 \pm 0.29$&$\mathbf{76.43 \pm 1.43}$&$\mathbf{2.37  \pm 0.01}$&$76.09 \pm 0.05$&$2.62 \pm 0.06$\\ \hline
CoWs\cite{demvsar2006statistical}       &$9$&$ $&$7$&$ $&$11$&$ $\\
\hline\hline
\end{tabular}\\
\footnotesize{Boldface denotes the best performance for each row. Accuracy: The higher is better; Time: The lower is better}\\
\end{table*}
\subsection{Model Selection in the Semi-Supervised Learning Paradigm}
Since there is no longer a requirement of data splitting under such limited labeled data, as a byproduct, LZO can be applied to more challenging tasks, e.g. Semi-supervised learning \cite{chapelle2009semi}. Thus, we further conduct experiments to investigate its flexibility. Since it is hard to execute the data splitting and model training with quite limited
labeled data, most of the existing semi-supervised learning works only report an empirical hyper-parameter without model selection \cite{2008Spectral,zhang2007semi,zhu2002learning,zhu2003semi,joachims1999transductive,kingma2014semi}. To this end, we adopt the Squared-loss Mutual Information Regularization (SMIR \footnote{\href{http://www.ms.k.u-tokyo.ac.jp/software/SMIR.zip}{http://www.ms.k.u-tokyo.ac.jp/software/SMIR.zip}.}) \cite{niu2013squared} as the base model, whose results are reported based on the 2-fold CV.
\subsubsection{Data Preparation}
We adopted eight publicly benchmark data-sets from a book \cite{chapelle2009semi} entitled Semi-Supervised Learning \footnote{\href{http://olivier.chapelle.cc/ssl-book/benchmarks.html}{http://olivier.chapelle.cc/ssl-book/benchmarks.html}.}) including g241c, g241n, Digit1, USPS, COIL2, BCI and Text. All data-sets are popular for benchmarking semi-supervised learning algorithms. The statistics of the adopted eight data-sets are listed in Table \ref{table4}.

\begin{table}[]
\centering%\normalsize
    \caption{Statistics of Benchmark Data-sets on the Semi-supervised Learning Setting}
    \label{table4}
\begin{tabular}{cccc}
\hline\hline
       & \#Class & \#Dimensions & \#Data \\ \hline
g241c  & 2       & 241          & 1500   \\
g241n  & 2       & 241          & 1500   \\
Digit1 & 2       & 241          & 1500   \\
USPS   & 2       & 241          & 1500   \\
COIL   & 2       & 241          & 1500   \\
COIL2  & 6       & 241          & 1500   \\
BCI    & 2       & 117          & 400    \\
Text   & 2       & 11960        & 1500   \\ \hline\hline
\end{tabular}
\end{table}
\subsubsection{Experimental Setting}
For fair comparison, we follow the same settings of theit original paper of SMIR \cite{niu2013squared}, which configuration set Q contains the hyper-parameters $\gamma \in \{10^{-7},10^{-3},10^{-1},10^{1},10^3\} $ and $\lambda\in \frac{\gamma c}{n} + \{10^{-10},10^{-8},10^{-6},10^{-4},10^{-2}\}$ of SMIR and the kernel width is the median of all pairwise distances times  $\{1/15,1/10,1/5,1/2,1\}$. We use $m=n$ and $m=10 \times n$ as the sample size of the augmented validation set. Here, we directly report the results from their original paper and use classification error as the measurement, which is opposite to the accuracy.
\subsubsection{Experimental Results and Analysis}
The classification error is shown in Table \ref{table5}. For each training set, we choose the regularized hyper-parameter $\gamma \in \{10^{-7},10^{-3},10^{-1},10^{-1},10^3\} $ , the hyper-parameter $\lambda\in \frac{\gamma c}{n} + \{10^{-10},10^{-8},10^{-6},10^{-4},10^{-2}\}$ of SMIR and the kernel width is the median of all pairwise distances times  $\{1/15,1/10,1/5,1/2,1\}$ on the training set, and evaluate the test errors for the chosen parameters on the test set. From Table \ref{table5}, we can easily observe that the test error of LZO significantly outperforms 2-fold CV on almost all data-sets except Text data-set, since LZO makes full use of the whole precious labeled data. Meanwhile, the variance of the classification error is also reduced when generating more augmented validation samples (i.e., $n=10 \times m$). Consequently, such results illustrate that the proposed LZO is sound and effective for model selection under the semi-supervised setting.
\begin{table}[]
\centering
%\normalsize
\caption{Mean with Standard Errors of Classification Error(\%) on Semi-supervised Learning Setting (SMIR)}
    \label{table5}
\begin{tabular}{cccc}
\hline\hline
       & 2-fold & LZO ($m=n$) & LZO ($m=10\times n$) \\ \hline
g241c  &$31.69 \pm 0.66$&$\mathbf{27.31 \pm 2.74}$  &${27.56 \pm 0.62}$   \\
g241n  &$33.76 \pm 0.65$&$\mathbf{28.04 \pm 3.99}$  &${28.94 \pm 0.57}$   \\
Digit1 &$10.23 \pm 0.40$&$\mathbf{5.25 \pm 1.45}$   &${5.41 \pm 0.46}$  \\
USPS   &$12.23 \pm 0.40$&$\mathbf{7.67 \pm 1.61}$   &${8.17 \pm 0.69}$  \\
COIL   &$33.62 \pm 0.82$&$\mathbf{23.49 \pm 4.74}$  &${25.43 \pm 0.96}$   \\
COIL2  &$24.12 \pm 0.69$&$\mathbf{12.78 \pm 2.36}$  &${13.23 \pm 0.67}$ \\
Text   &$\mathbf{38.80 \pm 0.64}$&$38.89 \pm 3.80$  &${38.81 \pm 0.93}$   \\ \hline\hline
\end{tabular}\\
\footnotesize{Boldface denotes the best performance for each row. \\
Classification error (\%): The lower is better}\\
\end{table}
\section{Conclusions}
\label{section6}
%It should be noted that the proposed LZO is not limited to the 
%提供了一个新的范式。特别适合于
Model Selection is a perennial problem in the machine learning field. In this paper, we develop a novel validation approach named LZO based on the auxiliary/augmented validation set. Also, we provide a theoretical upper bound of the estimation bias of the proposed LZO and derive a mild principle for data augmentation. The experimental results show that the proposed LZO has high computational efficiency, effective performance and wide application prospects. More importantly, such a methodology is general and can likewise be adapted to more realistic learning paradigms such as online learning, unsupervised learning, self-supervised learning, active learning by designing the corresponding loss or measurement function for model selection. Therefore, in the future we plan to work more for further validating the flexibility of the proposed LZO in the wild range of learning paradigms.

\section*{Appendice}
\setcounter{theorem}{0}   %从零开始编号
\setcounter{corollary}{0}   %从零开始编号

%定义编号格式，在数字序号前加字符“A"

%\renewcommand{\thetable}{A\arabic{table}}
%\section{Software packages}
\begin{theorem}
    \label{sthe1}
     Let $G\in\mathcal{H}$ be the model or the hypothesis learned from the input data $\mathcal{D}$ by configuration $\mathit{f}$ (i.e.,$G=\mathit{f}(\mathcal{D})$). Let $\mathbb{P}$ and $\mathbb{Q}$ be the distribution of training data and augmented validation data, respectively. Then, for any $\delta \in (0,1)$, with probability at least $ 1-\delta$, the following holds:
\end{theorem}
\begin{equation}
\begin{aligned}
\mid \mathcal{L}_{\mathbb{P}}(G) -\mathcal{L}(G, v) \mid\leq & \frac{B}{\sqrt{2}} \sqrt{D_{J S}(\mathbb{P} \| \mathbb{Q})} \\
+&\sqrt{\frac{4}{m}\left(v \ln \frac{2 e m}{v}+\ln \frac{4}{\delta}\right)}
\end{aligned}
\end{equation}
\noindent
\textbf{Proof} Let $ \mathcal{L}_{\mathbb{Q}}(G)$ be the expected risk over distribution $\mathbb{Q}$, according to the Cauchy-Schwarz inequality, we easily have:
\begin{equation}
\begin{aligned}
\mid\mathcal{L}_{\mathbb{P}}(G)&-\mathcal{L}(G,\mathcal{V})\mid\\ 
&=\left|\mathcal{L}_{\mathbb{P}}(G)-\mathcal{L}_{\mathbb{Q}}(G)+\mathcal{L}_{\mathbb{Q}}(G)-\mathcal{L}(G, \mathcal{V})\right| \\
& \leq\left|\mathcal{L}_{\mathbb{P}}(G)-\mathcal{L}_{\mathbb{Q}}(G)\right|+\left|\mathcal{L}_{\mathbb{Q}}(G)-\mathcal{L}(G, \mathcal{V})\right|
\end{aligned}
\end{equation}
For the first term $\left|\mathcal{L}_{\mathbb{P}}(G)-\mathcal{L}_{\mathbb{Q}}(G)\right|$, according to the \textbf{Theorem} 1 in Shui,  \MakeLowercase{\textit{et al.}} \cite{shui2020beyond}, we have:
\begin{equation}
\left|\mathcal{L}_{\mathbb{P}}(G)-\mathcal{L}_{\mathbb{Q}}(G)\right| \leq \frac{B}{\sqrt{2}} \sqrt{D_{J S}(\mathbb{P} \| \mathbb{Q})}
\end{equation}
For the second term $\left|\mathcal{L}_{\mathbb{Q}}(G)-\mathcal{L}(G, \mathcal{V}) \right|$, according to the Theorem 2 in Vapnik and Chervonenkis \cite{vapnik2015uniform}, we have:
\begin{equation}
\left|\mathcal{L}_{\mathbb{Q}}(G)-\mathcal{L}(G, \mathcal{V})\right| \leq \sqrt{\frac{4}{m}\left(d \ln \frac{2 e m}{d}+\ln \frac{4}{\delta}\right)}
\end{equation}
Thus, we easily have:
\begin{equation}
\begin{aligned}
\mid \mathcal{L}_{\mathbb{P}}(G) -\mathcal{L}(G, v) \mid\leq & \frac{B}{\sqrt{2}} \sqrt{D_{J S}(\mathbb{P} \| \mathbb{Q})} \\
+&\sqrt{\frac{4}{m}\left(v \ln \frac{2 e m}{v}+\ln \frac{4}{\delta}\right)}
\end{aligned}
\end{equation}
\noindent
\textbf{Q.E.D}
\begin{corollary}
\label{cols1}
The upper bound in Theorem \ref{the1} can be further decomposed as:
\end{corollary}
\begin{equation}
\begin{aligned}
\mid \mathcal{L}_{\mathbb{P}}(G) -\mathcal{L}(G, \mathcal{V}) \mid\leq & \frac{B}{\sqrt{2}} D \\
+&\sqrt{\frac{4}{m}\left(v \ln \frac{2 e m}{v}+\ln \frac{4}{\delta}\right)}
\end{aligned}
\end{equation}
where $D=\sqrt{\mathbb{E}_{y \sim \mathbb{P}(y)} D_{J S}(y \mid x)+\mathbb{E}_{y \sim Q(y)} D_{J S}(y \mid x)}+\sqrt{D_{J S}(y)}$. For convenience, we denote $D_{J S}(y \mid x)$ as $D_{J S}(\mathbb{P}(x \mid y) \| \mathbb{Q}(x \mid y))$ and $D_{J S}(y)$  as $D_{J S}(\mathbb{P}(y) \| \mathbb{Q}(y))$).
\noindent
\textbf{Proof:}
To proof \textbf{Corollary} \ref{cols1}, we first proof that $D_{K L}(\mathbb{P} \| \mathbb{M})=\mathbb{E}_{y \sim \mathbb{P}(y)} D_{K L}(\mathbb{P}(x \mid y) \| \mathbb{M}(x \mid y))+
D_{K L}(\mathbb{P}(y) \| \mathbb{M}(y)) .$ 
\begin{equation}
\label{eq12}
\begin{aligned}
D_{K L}&(\mathbb{P}\|\mathbb{M})\\
&=\int_{\mathcal{X} \times \mathcal{Y}} \mathbb{P}(x, y) \log \frac{\mathbb{P}(x \mid y) \mathbb{P}(y)}{\mathbb{M}(x \mid y) \mathbb{M}(y)} d x d y \\
& =\int_{\mathcal{Y}} \mathbb{P}(y)\left(\int_{\mathcal{X}} \mathbb{P}(\mathrm{x} \mid \mathrm{y}) \log \frac{\mathbb{P}(x \mid y)}{\mathbb{M}(x \mid y)} d x\right) d y \\
& +\left(\int_{\mathcal{X}} \mathbb{P}(\mathrm{x} \mid \mathrm{y}) d x\right)\left(\int_{\mathcal{Y}} \mathbb{P}(\mathrm{y}) \log \frac{\mathbb{P}(y)}{\mathbb{M}(y)} d x\right) \\
& =\mathbb{E}_{y \sim \mathbb{P}(y)} D_{K L}\left(\mathbb{P}(x \mid y)|| \mathbb{M}(x \mid y)\right)\\
&+ D_{K L}(\mathbb{P}(y)|| \mathbb{M}(y))\\
\end{aligned}
\end{equation}
Then, incorporating Eq. \ref{eq12} into \textbf{Definition} \ref{def1}, we have:
\begin{equation}
\begin{aligned}
D_{J S}&(\mathbb{P} \| \mathbb{Q})=\frac{1}{2}\left[D_{K L}(\mathbb{P} \| \mathbb{M})+D_{K L}(\mathbb{Q} \| \mathbb{M})\right] \\
&=\frac{1}{2}\left[\mathbb{E}_{y \sim \mathbb{P}(y)} D_{K L}(\mathbb{P}(x \mid y) \| \mathbb{M}(x \mid y))\right.\\
&+\mathbb{E}_{y \sim \mathbb{Q}(y)} D_{K L}(\mathbb{Q}(x \mid y) \| \mathbb{M}(x \mid y)) \\
&\left.+D_{K L}(\mathbb{P}(y) \| \mathbb{M}(y))+D_{K L}(\mathbb{Q}(y) \| \mathbb{M}(y))\right] \\
&\leq D_{J S}(y)+\left (\mathbb{E}_{y \sim \mathbb{P}(y)} D_{J S}(y \mid x)\right.\\
&\left.+\mathbb{E}_{y \sim \mathbb{Q}(y)} D_{J S}(y \mid x)\right)
\end{aligned}
\end{equation}
Next, according to the Cauchy-Schwartz inequality, we easily have:
\begin{equation}
\sqrt{D_{J S}(\mathbb{P} \| \mathbb{Q})} \leq D
\end{equation}
Finally, incorporating the upper inequality into \textbf{Theorem} \ref{the1}, we have:

\begin{equation}
\begin{aligned}
\mid \mathcal{L}_{\mathbb{P}}(G) -\mathcal{L}(G, \mathcal{V}) \mid\leq & \frac{B}{\sqrt{2}} D \\
+&\sqrt{\frac{4}{m}\left(v \ln \frac{2 e m}{v}+\ln \frac{4}{\delta}\right)}
\end{aligned}
\end{equation}
\noindent
\textbf{Q.E.D}
%
%% you can choose not to have a title for an appendix
%% if you want by leaving the argument blank
%%\section{}
%%Appendix two text goes here.
%
%
%
%\section{Related Definition}
%Extreme Value Theory\cite{kotz2000extreme}, also known as Fisher-Tippett Theorem, is a branch of statistics analyzing the distribution of data of abnormally high or low values. We here briefly review it as follows.
%\begin{thm}\label{thm1}[Extreme Value Theory\cite{kotz2000extreme}]
%Let $(v_1,v_2,...)$ be a sequence of i.i.d samples. Let $\zeta_n=\max\{v_1,...,v_n\}$. If a sequence of pairs of real numbers $(a_n,b_n)$ exists such that each $a_n>0$ and $\lim_{z\rightarrow\infty}P(\frac{\zeta_n-b_n}{a_n}\leq z)=F(z)$ then if $F$ is a non-degenerate distribution function, it belongs to the Gumbel, the Fr$\acute{e}$chet or the Reversed Weibull family.
%\end{thm}

%\begin{myDef}[Universum Data\cite{weston2006inference}]
%The universum data usually denotes the samples that do not belong to either class of interest for the specific classification problem.
%
%\end{myDef}

\ifCLASSOPTIONcompsoc
  \section*{Acknowledgments}
\else
  \section*{Acknowledgment}
\fi

The authors would like to thank Prof. Hui Xue, Prof. Yunyun Wang, Prof. Lishan Qiao, Dr. Yunxia Lin, Dr. Zirui Wang and Dr. Jiexi Liu for correcting the English language usage, grammar, punctuation, and spelling. This work is supported in part by Key Program of NSFC under Grant No. 61732006 and the NSFC under Grant No. 62076124.

\ifCLASSOPTIONcaptionsoff
  \newpage
\fi



%\bibliographystyle{IEEEtran}
%\begin{thebibliography}{1}
%\end{thebibliography}

\bibliographystyle{ieeetr}
\bibliography{mybibfile}

%%% Authors information
%\end{IEEEbiography}
%% insert where needed to balance the two columns on the last page with
%\newpage
\newpage
\begin{IEEEbiography}[{\includegraphics[width=1in,height=1.25in,clip,keepaspectratio]{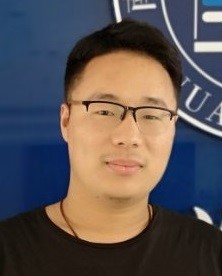}}]{Weikai Li}
received his B.S. degree in Information and Computing Science from Chongqing Jiaotong University in 2015. In 2018, he completed his M.S. degree in computer science and technique at Chongqing Jiaotong University. He is currently pursuing the Ph.D. degree with the College of Computer Science \& Technology, Nanjing University of Aeronautics and Astronautics. His research interests include pattern recognition and machine learning..
\end{IEEEbiography}
\begin{IEEEbiography}[{\includegraphics[width=1in,height=1.25in,clip,keepaspectratio]{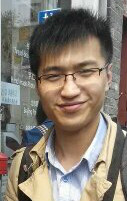}}]{Chuanxing Geng}
received the B.S. degree in mathematics from Liaocheng University in 2013. In 2016, he completed his M.S. degree in applied mathematics at Ningbo University. In 2020, he received a Ph.D. degree in computer science and technology at Nanjing University of Aeronautics and Astronautics. His research interests include pattern recognition and machine learning.
\end{IEEEbiography}
\begin{IEEEbiography}[{\includegraphics[width=1in,height=1.25in,clip,keepaspectratio]{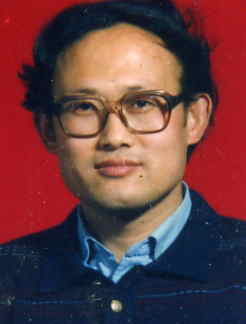}}]{Songcan Chen} 
received his B.S. degree in mathematics from Hangzhou University (now merged into Zhejiang University) in 1983. In 1985, he completed his M.S. degree in computer applications at Shanghai Jiaotong University and then worked at NUAA in January 1986. There he received a Ph.D. degree in communication and information systems in 1997. Since 1998, as a full-time professor, he has been with the College of Computer Science \& Technology at NUAA. His research interests include pattern recognition, machine learning and neural computing. He is also an IAPR Fellow.
\end{IEEEbiography}

%
%\begin{IEEEbiography}[{\includegraphics[width=1.2in,height=1.25in,clip,keepaspectratio]{huangsj_face}}]{Sheng-Jun Huang}
%received the BSc and PhD degrees in computer science from Nanjing University, China, in 2008 and 2014, respectively.
%He joined the College of Computer Science and Technology at Nanjing University of Aeronautics and Astronautics as an assistant professor in 2014.
%His main research interests include machine learning and patter recognition.
%He has won the Microsoft Fellowship Award (2011) and the Best Poster Award at KDD'12.
%He has been a Program Committee member of several conferences including KDD¡¯15, IJCAI¡¯15, ICDM¡¯15, etc.
%\end{IEEEbiography}
%
%


%\vfill


\end{document}